\documentclass{article}
\usepackage{spconf,amsmath,graphicx,url}

\usepackage{algorithm,algpseudocode,listings,xcolor}
\title{Spoken English Intelligibility Remediation with PocketSphinx
	Alignment and Feature Extraction Improves Substantially over 
	the State of the Art}

\name{Yuan Gao\thanks{$^1$ gyfreedom93@163.com (corresponding
		author for inquiries in Chinese)}$^{2}$ \qquad Brij Mohan Lal Srivastava\thanks{$^2$ 
		contactbrijmohan@gmail.com
		(corresponding author for inquiries in Indic languages)}$^{1}$ \qquad James Salsman\thanks{$^3$ jim@talknicer.com (corresponding 
		author for inquiries in other languages)}$^{3 4}$}

\address{$^{1}$ International Institute of Information Technology, Hyderabad, India \\
    $^{2}$ Beijing University of Technology,
    $^{3}$ 17zuoye.com,
    $^{4}$ TalkNicer.com, LLC}

\begin{document}
%
\maketitle
\begin{abstract}
We use automatic speech recognition to assess spoken English 
learner pronunciation based on the authentic intelligibility 
of the learners' spoken responses determined from support vector 
machine (SVM) classifier or deep learning neural network model 
predictions of transcription correctness. 
Using numeric features produced by PocketSphinx alignment mode and
many recognition passes searching for the substitution and 
deletion of each expected phoneme and insertion of unexpected phonemes
in sequence, the SVM models achieve 82\% agreement with the accuracy
of Amazon Mechanical Turk crowdworker transcriptions, up from 75\% 
reported by multiple independent researchers.
Using such features with SVM classifier probability prediction 
models can help computer-aided pronunciation teaching (CAPT) 
systems provide intelligibility remediation.
\end{abstract}

\section{Introduction}

\textbf{Authentic intelligibility}, the ability of listeners to
correctly transcribe recorded utterances, initially used for CAPT 
by \cite{Kib2011} and \cite{KHN2014}, 
is a better measure of pronunciation assessment for spoken 
language learners compared to mispronunciations identified by expert 
pronunciation judges or panels of experts, because such
mispronunciations are associated with only 16\% of intelligibility
problems, according to \cite{Lou2015}, who state:

\begin{quote}
We investigated ... which words are likely to be misrecognized 
and which words are likely to be marked as pronunciation errors. 
We found that only 16\% of the variability in word-level 
intelligibility can be explained by the presence of obvious 
mispronunciations.  Words perceived as mispronounced remain intelligible in about half of all cases. At the same time ... annotators were often
unable to identify the word when listening to the audio but did not
perceive it as mispronounced when presented with its transcription.
\end{quote}

This substantial improvement is not yet well understood by most 
CAPT community. 
Currently, expert human pronunciation judges assess student 
performance, often with large inter-rater variability between 
experts scoring the same utterances. Since most 
formal mispronunciations do not substantially impede understanding 
of spoken language,  automatic speech recognition CAPT systems 
trained to approximate the subjective assessments of judges do 
not perform as well as might be expected after intensive
work on the issue by several hundred researchers spanning decades
(\cite{ChenLi2016}, \cite{JLBIB}.)
While there are many commercial CAPT applications, there is no 
consensus among speech language pathologists about which of them,
if any, work well (\cite{Dudy2017}).

In high stakes situations, systems imitating subjective assessments 
of human judges have, for example, prevented native English speakers 
and trained English language radio announcers from immigrating to
Australia (\cite{Ferrier2017}, \cite{AAP2017}).
A more technical related problem with traditional CAPT approaches 
is that popular pronunciation assessment metrics, primarily  
\textbf{goodness of pronunciation} (GOP) as defined by
\cite{Witt2000}, are quotients with such vaguely specified
denominators \cite{Qian2015} that they tend to correlate weakly 
with authentic intelligibility. Earlier work suffers from similar
problems.

We are offering remediation of authentic intelligibility
for English CAPT to 17zuoye.com's 30 million K-6 English 
language students in China, and we are deploying the same technology 
in the Wikimedia Foundation's \textit{Wiktionary} dictionaries along 
with their phonetics and pronunciation articles in Wikipedia 
to provide free CAPT assessment and remediation exercises.
We are measuring which feedback choices perform the best for 
student proficiency outcomes, and studying the possibility of using 
students to provide transcriptions instead of paid crowdworkers.

\vspace{-1em}
\section{Adapting PocketSphinx for feature extraction}
\label{sec:adapting}

We chose to use PocketSphinx\cite{DHD2006} system's alignment 
routines.  We tried a two-pass alignment approach over a fixed 
grammar by using the time endpoints from recognizing the phonemes 
of  the expected utterance in sequence, using a finite state grammar
with no alternative or optional components other than silence, 
defined using a \textit{JSpeech Grammar Format} file. The 
results for the first pass were discarded, because its purpose 
was solely to perform \textit{cepstral mean normalization} 
for adapting to  the audio characteristics of the microphone, channel, 
and noise.  We found that grammar-based alignment, which is optimized
for speed instead of accuracy, resulted in less correctly predictive
features than using a single pass of the alignment API functions, 
which are only available from the PocketSphinx C API instead of 
command line invocations. 

\begin{figure}
	\includegraphics[width=\linewidth]{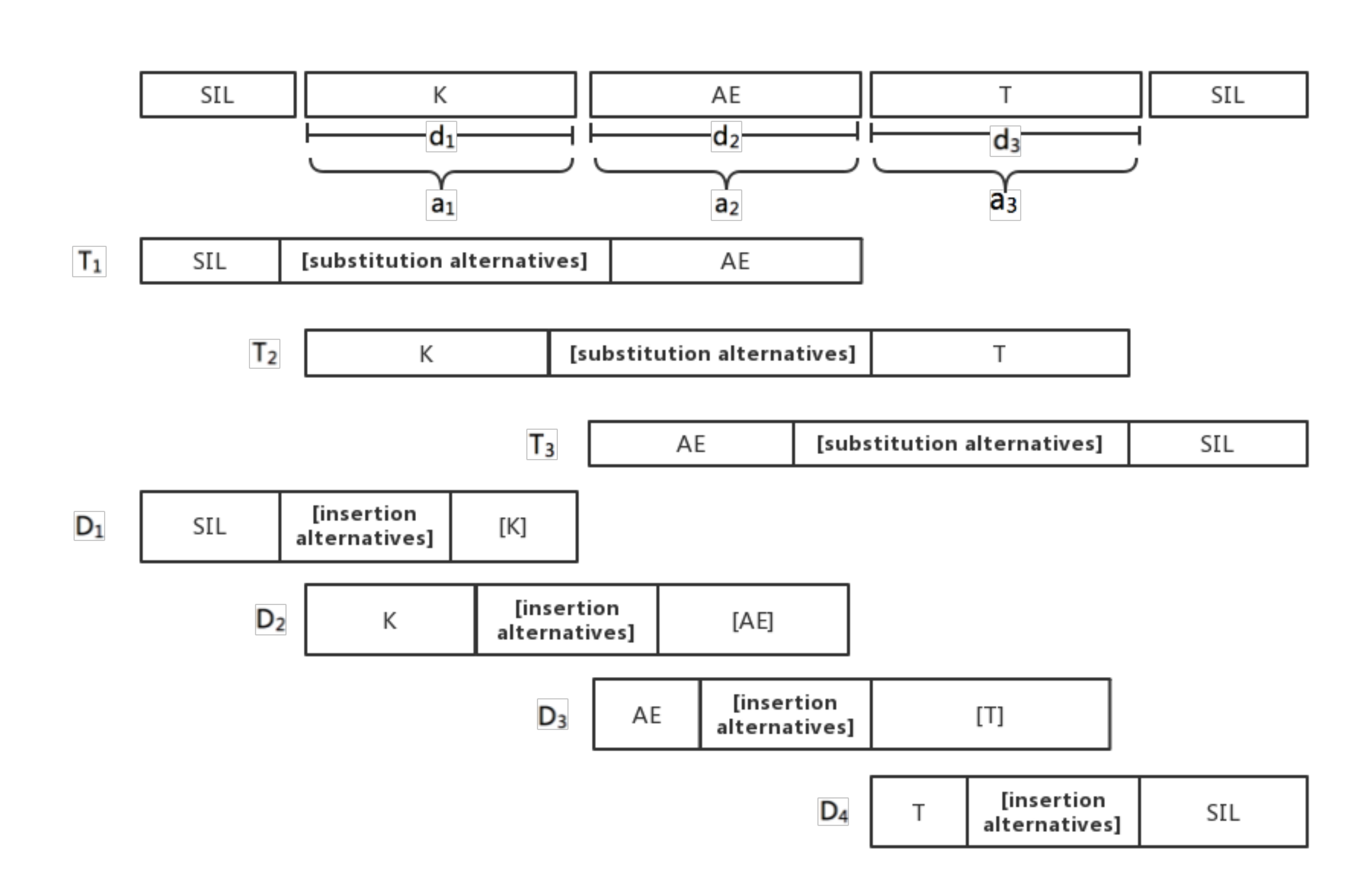}
	\caption{Feature extraction: The three phonemes of 
	the word `cat' are aligned, producing durations $ d_n $ and 
	acoustic scores $ a_n $. Then several passes of recognition 
	to the audio aligned to groups of three ($ T_n $) and two 
	($ D_n $) phonemes are used to measure phoneme 
	substitutions, and insertions and deletions, respectively.}
    \label{featex_diag}
\end{figure}

The results of the alignment are used to select audio sub-segments 
of the utterance to indicate substitutions of expected phonemes, 
insertions of unexpected phonemes, deletions of the expected phonemes,
and five physiological measures of the vocal tract, in multiple 
subsequent recognizer passes of each three and two adjacent phonemes 
at a time. Figure~\ref{featex_diag} illustrates the non-physiological 
part of this feature extraction process.

After alignment, we run the recognizer on each sub-segment of the audio corresponding to each three aligned phonemes in sequence, and count how soon the expected phoneme occurs in the n-best recognition results. Then we run the recognizer on each sub-segment corresponding to each two adjacent phonemes in sequence, simultaneously counting how frequently the initial expected phoneme is omitted when searching for the insertion of all 39 phonemes and silence in between the two expected phones.

The substitution detection pass focuses on three adjacent phonemes at
a time as located by the alignment routine. For the audio sub-segment
of each three adjacent phonemes from the alignment, we use a grammar
specifying the first and last of the three as the only options on
the ends, with an alternative allowing for any one phoneme (including
diphthongs) in the middle. The score, in the range $[0, 1]$, represents how high the expected middle phoneme ranks in the n-best results of all the possible phonemes in between the other two. We ask the recognizer for as many n-best results as possible, because sometimes a truncated grammar result (e.g., only two phonemes instead of three) result, but we often get at least 30 results from the 40 possible phonemes and silence, and sometimes get 70 results. The insertion and deletion pass operates on the audio sub-segments of two adjacent phonemes at a time, using a grammar to look for the first expected phoneme in the front as the only possibility, followed by an optional alternative of any phoneme other than the expected second phoneme counting as insertions, and then followed by the expected second phoneme specified as optional to account for deletion. Each time an insertion or deletion is returned in the n-best results before only the expected two phonemes are returned, the $[0, 1]$ score is reduced.

We also produce each phoneme's duration and the logarithm of its 
acoustic score from the alignment phase as features in our SVM 
or DNN classifier feature inputs. For each phoneme, we produce: 
(1) a duration; (2) an acoustic score from the alignment, corresponding 
to the numerator of the GOP score of \cite{Witt2000}; 
(3) a $[0, 1]$ score measuring phoneme substitution, and 
(4) a $[0, 1]$ score measuring insertions and deletions.
One final additional insertion and deletion measurement appears at the 
end of the feature vector for each word; in a multi-word phrase, 
that final score is shared as identical to the first insertion and
deletion measurement of the next word.

As this article was going to press, we added five additional 
physiological features per phoneme, relating to \textbf{place, 
closedness, roundedness, voicing,} and the proportion of neighboring
phonemes less likely. (\cite{RSB2012})

We use some non-standard PocketSphinx parameters. We use a frame rate 
of 65 frames per second instead of 100, because learners are not 
likely  to speak very quickly. We use a \textit{-topn} value of 
64 instead of 2. This provides more accurate recognition results 
at the expense of longer runtime, but our feature extraction 
system runs in better than real time in a single thread of a 
2016 Apple MacBook Air, and on user's browsers as a 
\textit{pocketsphinx.js} adaptation in 
JavaScript.  We use a \textit{-beam} parameter of 
$ 10^{-57} $, a \textit{-wbeam} parameter of $ 10^{-56} $, and a 
\textit{-maxhmmpf} value of $ -1 $ for the same reason. We set 
\textit{-fsgusefiller} to "no" so that optional pauses are not 
assumed between every word, allowing us to define words comprised of 
a single CMUBET phoneme without slowdown.

\subsection{Compiling featex.c with PocketSphinx}

The C source code to perform the feature extraction, \textbf{featex.c}, 
and instructions for compiling and using it are available under the MIT 
open source code license at:
\\
\texttt{https://github.com/jsalsman/featex}

\section{Using pocketsphinx.js in web browsers}

Feature extraction can take place in web browsers' JavaScript code 
using the \textbf{Emscripten} system of compiling C to JavaScript, 
and audio recorded in web browsers supporting microphone input.
During the initialization process, the browser is checked for microphone 
availability and the sampling frequency at which it operates. A media 
source stream is requested to record audio from the microphone, and 
connected to a recorder thread which listens or stops listening based 
on browser user interface events. The pocketsphinx.js module is 
initialized inside a web worker to asynchronously call the alignment 
and feature extraction modules.

\begin{algorithm}
	\begin{algorithmic}[1]
		\caption{Web client algorithm}
        \State The user presses the 'Record' button.
        \State The recorder thread starts listening.
        \State The user presses the 'Stop' button.
        \State The recorded audio is converted and 
        downsampled if necessary.
        \State The extracted feature vector and word is sent 
        to the intelligibility prediction service (see sections 5.1 and 7.) 
        \State Assessment feedback is provided to the user.
	\end{algorithmic}
\end{algorithm}

The integrated code and detailed compilation instructions can be 
found at \cite{psjsbrij2017}. For more information and an example of an integrated web browser system, please see \cite{iremedybrij2017}. For an example of how such a system might be integrated into Wiktionary, please see \cite{slbrij2017}.

\section{Obtaining transcriptions of student utterances}


We consistently obtained faster responses from Amazon 
Mechanical Turk when paying \$0.03 per transcript compared to 
\$0.15. We believe crowdworkers prefer to do low-paying tasks because 
they are likely to be easier and will cause fewer problems if the 
work is rejected. We are studying the possibility of using our English 
learners to provide transcriptions instead of paying crowdworkers, as 
\textit{bona fide} listening comprehension and typing exercises 
suitable for assessments in their own right.

\section{Predicting intelligibility}

\begin{figure}
	\includegraphics[width=\linewidth]{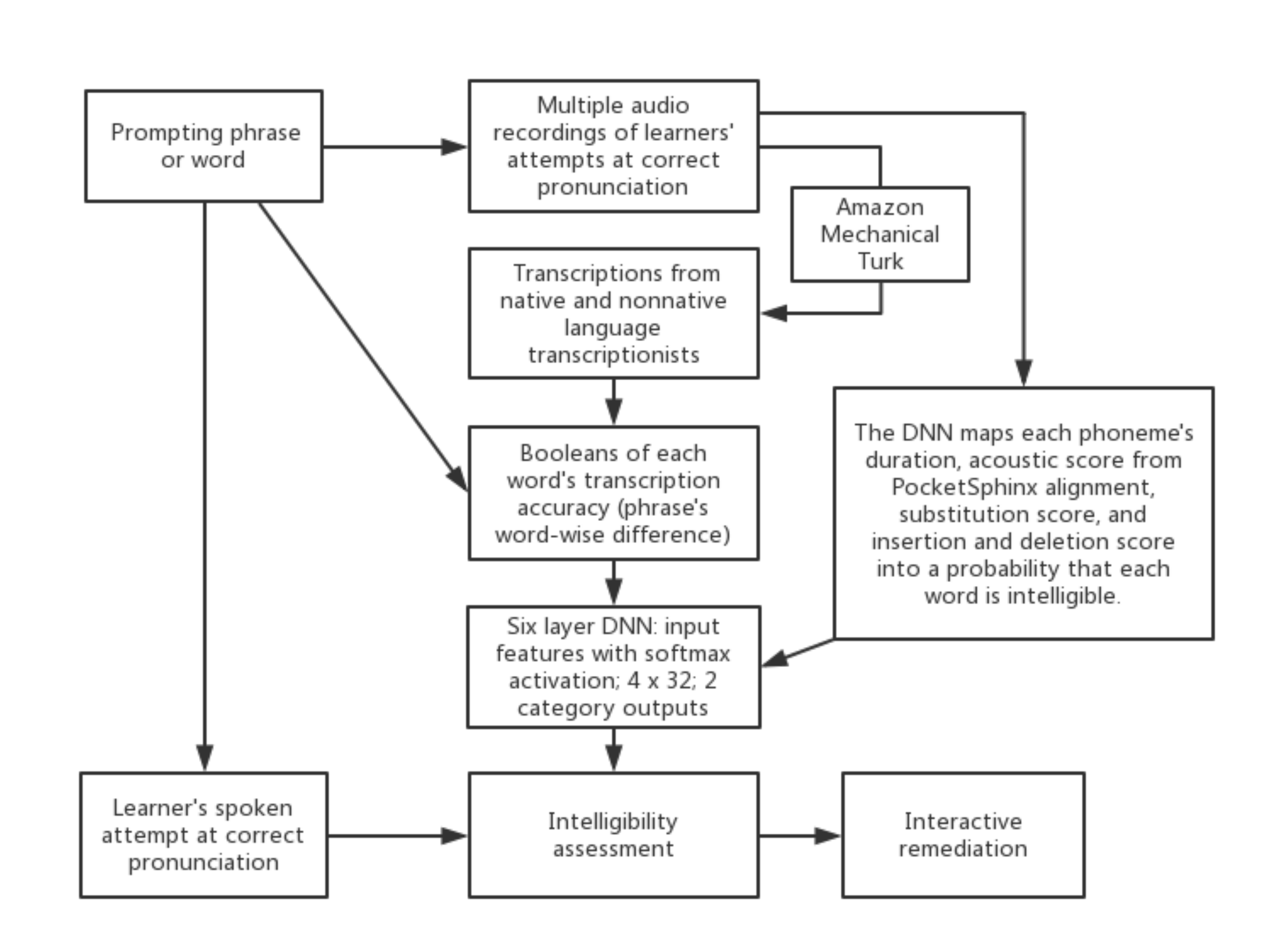}
	\caption{Predicting intelligibility.}
    \label{pred_intell}
\end{figure}

Using nine features per phoneme as described above (but not 
depicted) with support vector machine classification routines 
from the Python \textbf{Scikit-learn SVC} library configured
with a radial basis function kernel and probability prediction, 
we obtain 82\% accuracy in predicting the intelligibility of 
about 700 basic English words in agreement with Amazon Mechanical
Turk workers, using about 30 recordings per words and four 
transcripts per recording. We have measured strong evidence 
that increasing the number of recordings per word and transcripts 
per recording can result in very substantial accuracy improvements. 
We have obtained similar results on longer phrases. 
Using the four features per phoneme to train 
a \textit{linear logistic regression} model, we only get 
75\% accuracy, which was reported by \cite{Kib2011} and 
\cite{KHN2014} and the ETS (\cite{Lou2015}). 

For a client-server system to predict word intelligibility from 
feature vectors, please see \cite{pronevalbrij2017}.

\section{Measuring the accuracy of intelligibility assessment}

When different transcripts of the same utterance of a word show
both intelligible and unintelligible results, 
we measure accuracy as a fraction of the best 
possible result. For example, if the same utterance was transcribed 
correctly by three transcriptionists but incorrectly by a fourth, the 
maximum unadjusted accuracy achievable from predicting that 
utterance's intelligibility is 75\%, so an unadjusted accuracy of 50\% 
is adjusted to be 67\%, representing the proportion of the maximum 
possible accuracy. In practice, the probability of intelligibility is a 
floating point value in [0, 1], which is typically compared to a 
threshold, the estimated intelligibility of other words in the same 
phrase, or both, so the accuracy with which we can predict 
intelligibility by transcriptionists is used as a benchmark by which 
we can measure the relative utility of different prediction methods.

\vspace{-1em}
\section{Determining optimal feedback}

We use the modeled probability of intelligibility of each word in 
a prompt word or phrase to help students improve their pronunciation 
by providing audiovisual feedback indicating which word(s) were
pronounced the worst. How many words to indicate were not pronounced 
well after each utterance is an open question.

\begin{figure}
	\includegraphics[width=\linewidth]{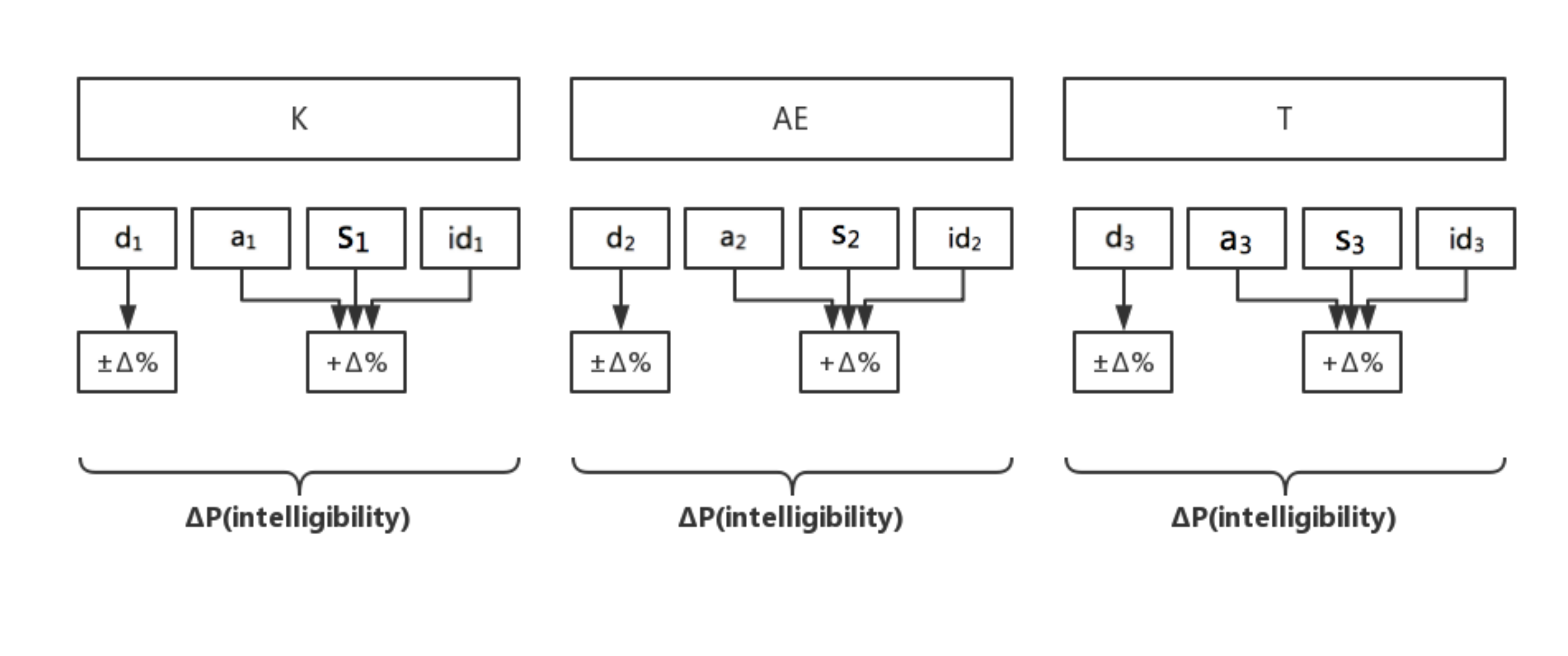}
	\caption{Determining feedback: Adjusting the feature
	scores for each phoneme changes the probability of intelligibility
	of the whole word. The adjustments which make the best changes 
	signal which phoneme(s) need improvement the most.}
    \label{fig:feedback}
\end{figure}

For words which are not considered sufficiently intelligible, we can 
use the SVM classifier probability prediction models 
to determine which identical numerical improvement 
to each phoneme's non-duration features improves the probability of 
word intelligibility the most. We can also see how increasing and  
decreasing each phoneme's duration improves the intelligibility of 
the word. Such adjustments to the features derived from automatic 
speech recognition may be more useful as products than sums to
identify the specific phoneme(s) most in need of improvement in 
the less unintelligible word(s). Figure~\ref{fig:feedback} shows how we determine the phoneme-level feedback for each word.

\section{Conclusion}

Using PocketSphinx automatic speech recognition with improved 
phonetic accuracy features training SVM prediction models 
can help CAPT systems provide better intelligibility remediation. Researchers and commercial software publishers should try to 
understand the reasons this technique is superior to the state 
of the art, and adopt it for improved CAPT outcomes. 


\section{acknowledgments}
	We thank 17zuoye.com (China), Zzish.com (UK), 
    Prof. Seiichi Nakagawa,
	the Google Open Source Programs Office, 
    and the Wikimedia Foundation 
	for their kind financial support, suggestions, 
    personnel resources, and educational infrastructure.


\vfill\pagebreak

\bibliographystyle{IEEEbib}
\bibliography{strings,refs}

\begin{thebibliography}{10}

\bibitem{Kib2011}
Hiroshi Kibishi and Seiichi Nakagawa,
\newblock ``New feature parameters for pronunciation evaluation in english
  presentations at international conferences,''
\newblock in {\em Twelfth Annual Conference of the International Speech
  Communication Association}, 2011.

\bibitem{KHN2014}
Hiroshi Kibishi, Kuniaki Hirabayashi, and Seiichi Nakagawa,
\newblock ``A statistical method of evaluating the pronunciation
  proficiency/intelligibility of english presentations by japanese speakers,''
\newblock {\em ReCALL}, vol. 27, no. 1, pp. 58--83, 2015.

\bibitem{Lou2015}
Anastassia Loukina, Melissa Lopez, Keelan Evanini, David Suendermann-Oeft,
  Alexei~V Ivanov, and Klaus Zechner,
\newblock ``Pronunciation accuracy and intelligibility of non-native speech,''
\newblock in {\em Sixteenth Annual Conference of the International Speech
  Communication Association}, 2015.

\bibitem{ChenLi2016}
Nancy~F Chen and Haizhou Li,
\newblock ``Computer-assisted pronunciation training: From pronunciation
  scoring towards spoken language learning,''
\newblock in {\em Signal and Information Processing Association Annual Summit
  and Conference (APSIPA), 2016 Asia-Pacific}. IEEE, 2016, pp. 1--7.

\bibitem{JLBIB}
J.~2016. Llisterri,
\newblock ``Computer-assisted pronunciation teaching bibliography.,'' October
  2016,
\newblock [Online; posted 25-October-2016].

\bibitem{Dudy2017}
Shiran Dudy, Meysam Asgari, and Alexander Kain,
\newblock ``Pronunciation analysis for children with speech sound disorders,''
\newblock in {\em Engineering in Medicine and Biology Society (EMBC), 2015 37th
  Annual International Conference of the IEEE}. IEEE, 2015, pp. 5573--5576.

\bibitem{Ferrier2017}
T.~2017. Ferrier,
\newblock ``Australian ex-news reader with {English} degree fails robot's
  {English} test.,'' August 2017,
\newblock [Online; posted 9-August-2017].

\bibitem{AAP2017}
Australian Associated~Press. 2017.,
\newblock ``Computer says no: {Irish} vet fails oral {English} test needed to
  stay in {Australia},'' August 2017,
\newblock [Online; posted 8-August-2017].

\bibitem{Witt2000}
Silke~M Witt and Steve~J Young,
\newblock ``Phone-level pronunciation scoring and assessment for interactive
  language learning,''
\newblock {\em Speech communication}, vol. 30, no. 2, pp. 95--108, 2000.

\bibitem{Qian2015}
Xiaojun Qian, Helen Meng, and Frank Soong,
\newblock ``A two-pass framework of mispronunciation detection and diagnosis
  for computer-aided pronunciation training,''
\newblock {\em IEEE/ACM Transactions on Audio, Speech and Language Processing
  (TASLP)}, vol. 24, no. 6, pp. 1020--1028, 2016.

\bibitem{DHD2006}
David Huggins-Daines, Mohit Kumar, Arthur Chan, Alan~W Black, Mosur
  Ravishankar, and Alexander~I Rudnicky,
\newblock ``Pocketsphinx: A free, real-time continuous speech recognition
  system for hand-held devices,''
\newblock in {\em Acoustics, Speech and Signal Processing, 2006. ICASSP 2006
  Proceedings. 2006 IEEE International Conference on}. IEEE, 2006, vol.~1, pp.
  I--I.

\bibitem{RSB2012}
Srikanth Ronanki, James Salsman, and Li~Bo,
\newblock ``Pronunciation evaluation and mispronunciation detection using cmu
  sphinx,''
\newblock in {\em Proceedings of the 24th International Conference on
  Computational Linguistics (Mumbai, India) Workshop on Speech and Language
  Processing Tools in Education}. COLING, 2012, pp. 61--68.

\bibitem{psjsbrij2017}
``{Pocketsphinx.js},'' \url{https://github.com/brijmohan/pocketsphinx.js},
  2017.

\bibitem{iremedybrij2017}
``{Iremedy},'' \url{https://github.com/brijmohan/iremedy}, 2017.

\bibitem{slbrij2017}
``{Live demo},'' \url{https://brijmohan.github.io/iremedy/single_line.html},
  2017.

\bibitem{pronevalbrij2017}
``{Pronunciation evaluation service},''
  \url{https://github.com/brijmohan/proneval-service}, 2017.

\end{thebibliography}

\end{document}